\pdfoutput=1

\documentclass[11pt]{article}

\usepackage[preprint]{coling}

\usepackage{times}
\usepackage{latexsym}

\usepackage[T1]{fontenc}

\usepackage[utf8]{inputenc}

\usepackage{microtype}

\usepackage{inconsolata}

\usepackage{graphicx}

\usepackage{multirow}
\usepackage{arydshln}
\usepackage{amsmath}
\usepackage{algorithm}
\usepackage[noend]{algpseudocode}

\usepackage{booktabs}
\usepackage{enumitem}
\usepackage{amssymb}
\usepackage{pifont}

\definecolor{c1}{HTML}{AF794B} 
\definecolor{c2}{HTML}{5C6954} 
\definecolor{c3}{HTML}{335E7D} 
\definecolor{c4}{HTML}{CD8A70}
\definecolor{c5}{HTML}{ADCACB}
\definecolor{gray}{HTML}{AAABA8}

\title{Extract-and-Abstract: Unifying Extractive and Abstractive Summarization within Single Encoder-Decoder Framework}

\author{Yuping Wu, Hao Li, Goran Nenadic, Xiao-Jun Zeng\thanks{Correpsonding Author.} \\
        Department of Computer Science \\ University of Manchester \\
        \texttt{\{yuping.wu, hao.li-2, gnenadic, x.zeng\}@manchester.ac.uk}
        }

\begin{document}
\maketitle
\begin{abstract}
Extract-then-Abstract is a naturally coherent paradigm to conduct abstractive summarization with the help of salient information identified by the extractive model. Previous works that adopt this paradigm train the extractor and abstractor separately and introduce extra parameters to highlight the extracted salients to the abstractor, which results in error accumulation and additional training costs. In this paper, we first introduce a parameter-free highlight method into the encoder-decoder framework: replacing the encoder attention mask with a saliency mask in the cross-attention module to force the decoder to focus only on salient parts of the input. A preliminary analysis compares different highlight methods, demonstrating the effectiveness of our saliency mask. We further propose the novel extract-and-abstract paradigm, $\textsc{ExtAbs}$\footnote{Codes available at \url{https://anonymous.4open.science/r/ExtAbs}.}, which jointly and seamlessly performs \textbf{Ext}ractive and \textbf{Abs}tractive summarization tasks within single encoder-decoder model to reduce error accumulation. In $\textsc{ExtAbs}$, the vanilla encoder is augmented to extract salients, and the vanilla decoder is modified with the proposed saliency mask to generate summaries. Built upon $\textsc{BART}$ and $\textsc{PEGASUS}$, experiments on three datasets show that $\textsc{ExtAbs}$ can achieve superior performance than baselines on the extractive task and performs comparable, or even better than the vanilla models on the abstractive task.
\end{abstract}

\section{Introduction}

The automatic text summarization task aims to condense the important information in a given text and form a summary. The extractive and abstractive are the two most common approaches to this task by extracting the most salient textual segments in the text or generating a sequence of words with salient information. 
With the great success achieved by Transformer, most of recent extractive and abstractive summarization models \citep{cheng-etal-2023-set, li-etal-2023-hear} are established from pre-trained Transformer-based models, among which, the encoder-decoder architecture dominates.

\begin{figure*}
  \centering
  \includegraphics[scale=0.5]{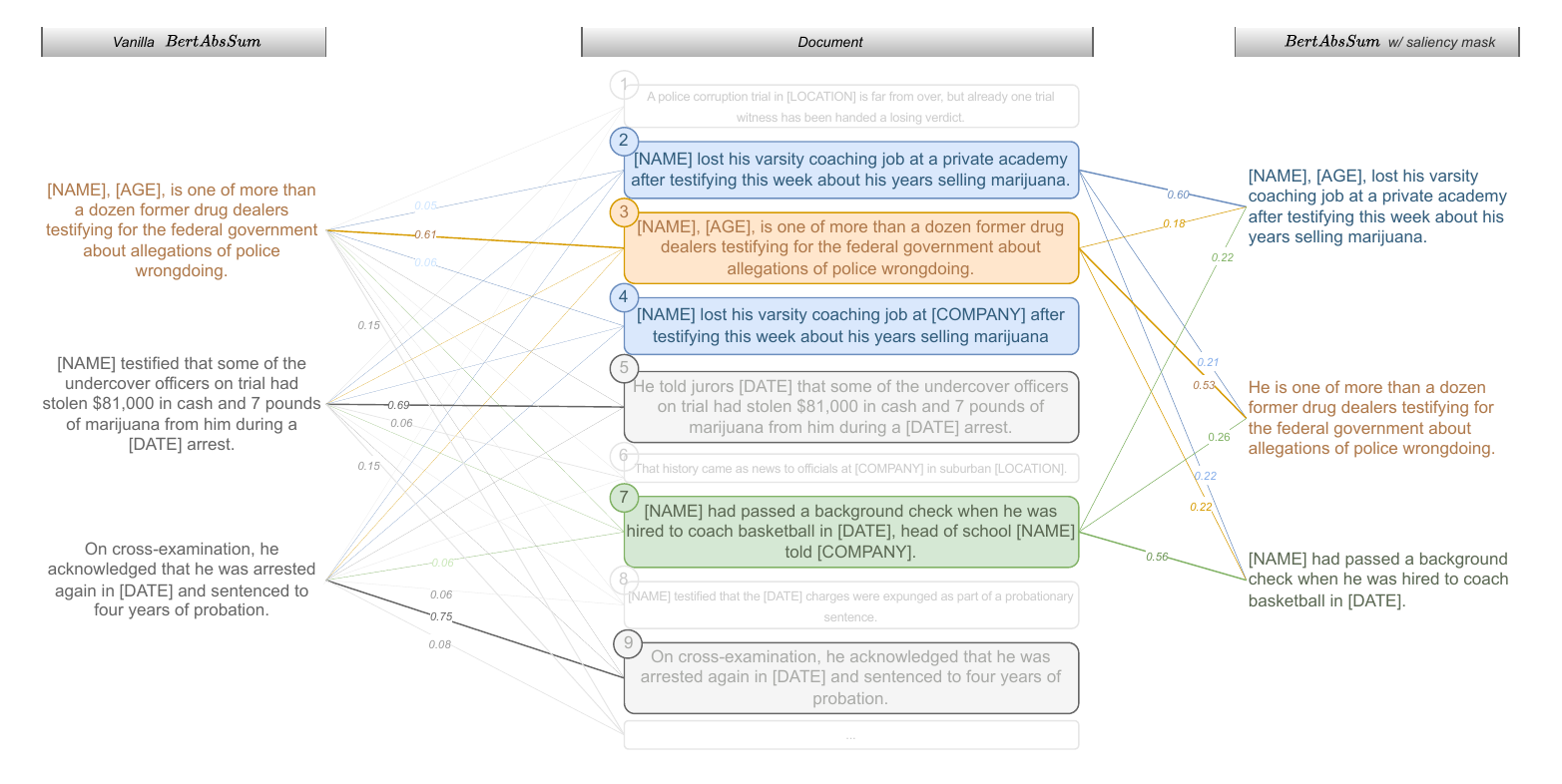}
  \caption{Simplified visualization of the cross-attention values of the first head at the penultimate decoder layer in $\textsc{BertAbsSum}$. The attention value between an input sentence and output sentence is calculated by summing token attention in an input sentence for each output token and averaging over the output sentence. Sentences \textcolor{c3}{\ding{173}}, \textcolor{c1}{\ding{174}} and \textcolor{c2}{\ding{178}} constitute \textbf{\textcolor{c3}{the} \textcolor{c1}{reference} \textcolor{c2}{summary}} and each colour represents a piece of salient information. }
  \label{Fig:example_demo}
\end{figure*}

The extract-then-abstract paradigm takes advantage of the inherent connection between the extractive and abstractive summarization by generating the abstractive summary with the utilization of extracted salient information. Existing works that explore this paradigm generate the summary with either the extractions as input only \citep{ernst-etal-2022-proposition, lebanoff-etal-2020-cascade} or the original input document with extractions highlighted \citep{bao2021contextualized, 10.1145/3477495.3531916, dou-etal-2021-gsum, adams-etal-2023-generating} as input. However, on the one hand, these works treat the extractor and abstractor as two functionally independent models which result in duplicate encoding, and most works train them individually, exposing the abstractor to errors accumulated from the extractor. On the other hand, the methods of highlighting extractions in these works inevitably introduce extra learning parameters and result in extra training costs, e.g., highlight embedding layer or extra encoder for extractions.

We first propose a parameter-free highlight method by augmenting attention, i.e., the saliency mask, a mask for salient tokens in the input sequence.
By replacing the vanilla encoder attention mask (i.e., the non-padded token mask) with the proposed saliency mask in the cross-attention module, the decoder is forced to only aggregate information from those salient tokens and ignore non-salient ones. 
As illustrated in Figure \ref{Fig:example_demo}, the vanilla model tends to be overconfident (sharper attention distribution) when determining the cross-attention values and thus fails to capture salient information with the wrong attention. Whereas with the proposed saliency mask, the model generates a more balanced attention distribution over the explicitly narrowed-down attentive scope.
To quantify the effectiveness of the saliency mask, we conduct preliminary analysis on the CNN/DM dataset to compare it with other highlight methods, and the results validate its effectiveness.

\begin{figure*}
  \centering
  \includegraphics[scale=0.45]{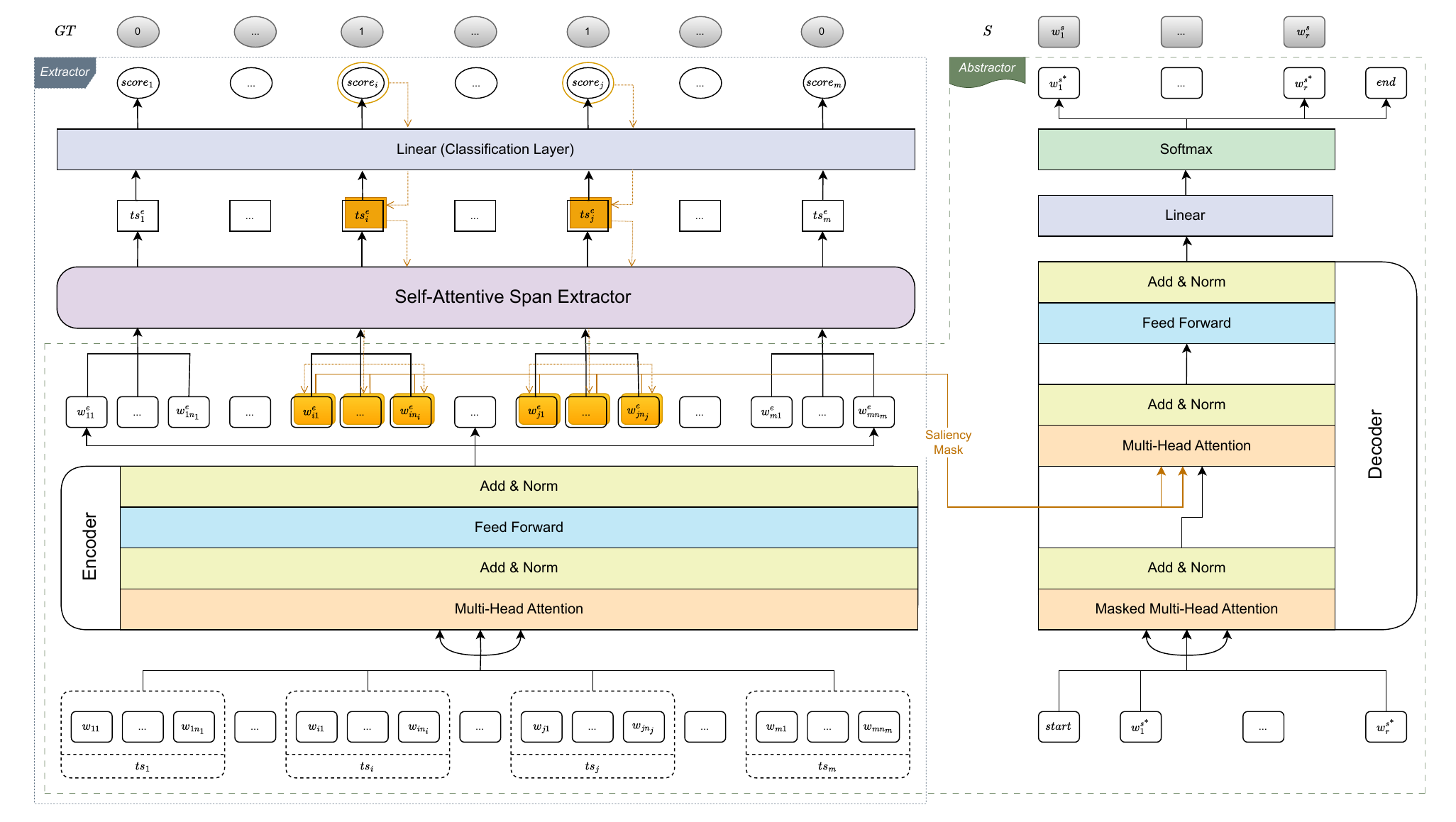}
  \caption{The architecture of the proposed $\textsc{ExtAbs}$. The left part serves as the extractor to perform saliency classification for each textual segment in the document. The right part is the abstractor, which generates a summary based on the encoder output and saliency mask determined by the predicted saliency scores.}
  \label{Fig:model}
\end{figure*}

Then, we propose the novel extract-and-abstract paradigm $\textsc{ExtAbs}$ (depicted in Figure \ref{Fig:model}), adapting any given encoder-decoder model to perform extractive and abstractive summarization jointly and seamlessly with encoder shared between the extractor and abstractor. In $\textsc{ExtAbs}$, the encoder is augmented by integrating the span extractor to learn text-span representations and a classification layer to perform the extractive classification task. The augmented encoder serves as the extractor to extract salient text spans. Together with the encoder, the decoder serves as the abstractor and is modified by alternating the encoder attention mask with the proposed saliency mask in the cross-attention module. The saliency mask is determined by the extractor's predicted top-$z$ salient text spans. Jointly training the augmented encoder and decoder enables the extract-and-abstract paradigm within a single encoder-decoder framework, removing the functional independence between the extractor and abstractor, along with duplicate encoding and error accumulation.
Experiments are conducted on the CNN/DM, Reddit and PubMed datasets with ROUGE scores and BARTScore as automatic metrics. On CNN/DM, $\textsc{ExtAbs}$ generates better abstractive summaries than the vanilla model. On Reddit and PubMed, it achieves the SOTA extractive performance while maintaining a comparable performance on the abstractive task compared with the vanilla model. Human evaluation also validates the quality of summaries generated by $\textsc{ExtAbs}$.

The contributions of this paper are threefold:
\begin{enumerate}[label=\arabic*)]
    \item We propose a parameter-free highlight method for the extract-then-abstract paradigm, i.e., the saliency mask, with preliminary analysis validating its effectiveness.
    \item We propose $\textsc{ExtAbs}$, a novel extract-and-abstract paradigm, which enhances the vanilla encoder-decoder model to jointly and seamlessly perform extractive and abstractive summarization. In $\textsc{ExtAbs}$, the jointly trained encoder not only mitigates errors that arise from disjoint processing for the abstractor but also improves extractive outputs (saliency masks). By learning from both extractive and abstractive instances and being optimized in a multi-task setting, the encoder achieves a higher standard of encoding performance, leading to better summarization performance.
    \item The experimental results show that $\textsc{ExtAbs}$ achieves superior abstractive performance than the vanilla model on CNN/DM and SOTA extractive performance on both Reddit and PubMed.
\end{enumerate}

\section{Related Work}

\paragraph{Text Summarization} Most previous works about automatic text summarization focus on doing it in only either an extractive or abstractive way. Along with the development of Transformer-based models, the paradigm of fine-tuning a pre-trained language model (PLM) dominates the methods in both ways. PLMs like $\textsc{BERT}$ have been widely adopted as the input encoder for extractive summarization models \citep{cheng-etal-2023-set, ruan-etal-2022-histruct, kwon-etal-2021-considering, zhong-etal-2020-extractive}. For abstractive summarization, PLMs such as $\textsc{BART}$ \cite{lewis-etal-2020-bart}, $\textsc{PEGASUS}$ \cite{10.5555/3524938.3525989}, ProphetNet \cite{qi-etal-2020-prophetnet} and Longformer \cite{DBLP:journals/corr/abs-2004-05150} were introduced for the better generation result. Most of the recent abstractive models \citep{pu-etal-2023-incorporating, goyal-etal-2022-hydrasum, liu-etal-2022-brio, dou-etal-2021-gsum} were built on one of these PLMs with the encoder-decoder architecture.


\paragraph{Extract-then-Abstract} Extracting salient information first and then performing abstractive summarization is naturally coherent. Some work \citep{bao2021contextualized, 10.1145/3477495.3531916, dou-etal-2021-gsum, ernst-etal-2022-proposition, lebanoff-etal-2020-cascade, adams-etal-2023-generating, pilault-etal-2020-extractive} explored this extract-then-abstract paradigm in a two-step manner, training the extractor and abstractor independently and using extra learning parameters to highlight extractions. Some work \citep{li-etal-2020-composing, song-etal-2022-improving} adapted the reinforcement learning to train the extractor by maximizing the reward derived from the abstractor. Only a few work \citep{hsu-etal-2018-unified, mendes-etal-2019-jointly} jointly trained the extractor and abstractor. Existing works essentially treat the extractor and abstractor as two functionally independent models.



Exploring the gaps of functional independence and additional highlight parameters in the existing extract-then-abstract paradigm, we develop a unified extract-and-abstract approach that trains the extractor and abstractor within a single encoder-decoder model. Building on the Transformer-based PLM, we introduce a saliency mask to highlight extracted salient information for the abstractor, resulting in an effective parameter-free method.

\section{Saliency Mask}
\label{sec:mask}
\paragraph{Motivation} The extract-then-abstract paradigm enhances the abstractive model by highlighting salient information identified by the extractive model, making it valuable to explore the effective highlight method.
Some previous works \citep{you-etal-2019-improving, xu-etal-2020-self, wang-etal-2022-salience} have explored different augmentation methods for the attention mechanism to emphasize tokens of interest such as salient tokens in the input sequence or control tokens. Their experimental results demonstrate the effectiveness of attention augmentation, inspiring us to propose a parameter-free highlight method by augmenting attention, i.e., saliency mask, the mask for salient tokens in the input sequence. Replacing the vanilla non-padded token mask with the saliency mask in the cross-attention module intuitively highlights salient information without introducing any extra parameters and explicitly forces the decoder to only attend to salient tokens.


\paragraph{Saliency Mask} Given an input sequence with $n$ number of tokens, let $l$ denote the list of indices of tokens in the salient parts of the input sequence, i.e., $l = [i, ..., j, ..., k]$ where $1 \leq i < j < k \leq n$. The saliency mask $mask \in \mathbb{R}^n$ is derived as follows:
\begin{gather} \label{equa:oracle_mask}
    mask_{i} = \begin{cases}
        1, & \text{if $i \in l$} \\
        0, & \text{otherwise}
    \end{cases}
\end{gather}

For a given encoder-decoder model, the cross-attention value between the transformed encoder output $K \in \mathbb{R}^{n \times d_k}$ and decoder intermediate output $Q \in \mathbb{R}^{m \times d_k}$ is modified to conduct the element-wise product with the saliency mask, i.e.,
\begin{gather}
  \tilde{attn}(Q,K) = softmax(\frac{QK^T}{\sqrt{d_k}}) \odot mask
  \label{Equa:attn}
\end{gather}
where $d_k$ is the dimension of $Q$ and $K$ and $m$ is the number of tokens in the decoded sequence. For the calculative consistency, the size of $mask$ is repeatedly expanded to be $\mathbb{R}^{m \times n}$.
As a result, the cross-attention module outputs
\begin{gather}
  \tilde{Attenion}(Q,K,V) = \tilde{attn}(Q,K)V
\end{gather}

\paragraph{Sentence-level and EDU-level Saliency Mask} Depending on the text granularity, the salient parts are not necessarily the same for a document. The two most common text granularities used in the summarization task are the sentence and sub-sentence, such as the Elementary Discourse Unit (EDU). EDU is defined as the terminal node in the Rhetorical Structure Theory (RST) \citep{Mann1988RhetoricalOrganization} and has been widely applied in summarization models \citep{pu-etal-2023-incorporating, adams-etal-2023-generating, wu-etal-2023-edu, xu-etal-2020-discourse}. It is evidenced by previous works that EDUs provide more fine-grained information than sentences for a summarizer.  Therefore, we propose EDU-level and sentence-level saliency masks derived from salient EDUs and sentences, respectively.

\paragraph{Preliminary Analysis} To verify the effectiveness of the proposed saliency mask, we conduct a preliminary analysis of the CNN/DM dataset to compare it with other highlight methods. The baseline highlight methods include $\textsc{ContextRewriter}$ \cite{bao2021contextualized} and $\textsc{EDURewriter}$ \cite{10.1145/3477495.3531916} which introduce additional group tag embedding layer for salient tokens, and $\textsc{GSum}$ \cite{dou-etal-2021-gsum} which incorporates an additional encoder for the salients. Following these previous works, a greedy selection algorithm is applied to determine the salients in a document by greedily maximizing the ROUGE scores between the selected salients and the reference summary (pseudo-code in Appendix \ref{appendix: greedy algo}). For a fair comparison, all methods are evaluated on $\textsc{BertSumAbs}$ \cite{liu-lapata-2019-text} with the pre-trained $\textsc{BERT}$-base as the encoder and 6 Transformer decoder layers as the decoder. Notably, $\textsc{EDURewriter}$ highlights EDU-level salients while the other two highlight at sentence level.

As shown in Table \ref{Tab:preliminary}, our proposed EDU-level saliency mask outperforms all three other highlight methods and the sentence-level saliency mask on ROUGE scores. Moreover, the saliency mask significantly outperforms the vanilla mask, where the whole input sequence is attended. The results demonstrate that our proposed highlight method better boosts abstractive model performance without introducing extra learning parameters.

\begin{table}[t]
  \centering
  \resizebox{\columnwidth}{!}
  {%
  \begin{tabular}{lccc}
    \hline
    \textbf{Method} & \textbf{ROUGE-1} & \textbf{ROUGE-2} & \textbf{ROUGE-L} \\
    \hline
    $\textsc{BertSumAbs}$ & 41.24 & 18.80 & 38.24 \\
    \cdashline{1-4}
    $\textsc{GSum}^{\dagger}$ & 55.18 & 32.54 & 52.06 \\
    $\textsc{ContextRewriter}^{\dagger}$ & 52.57 & 29.71 & 49.69 \\
    $\textsc{EDURewriter}^{\dagger}$ & 54.49 & 31.76 & 51.79 \\
    Saliency Mask (sentence) & 52.63 & 29.93 & 49.66 \\
    Saliency Mask (EDU) & \textbf{57.27} & \textbf{34.22} & \textbf{54.23} \\
    \hline
  \end{tabular} %
  }
  \caption{Results of different salient information highlight methods for model $\textsc{BertSumAbs}$. F1 scores are reported. $^{\dagger}$ indicates that the results are copied from the corresponding original paper.}
  \label{Tab:preliminary}
\end{table}


\section{$\textsc{ExtAbs}$}
\subsection{Problem Formulation}

Given a document $D$ consisting of $m$ textual segments and the $i$-th textual segment contains $n_i$ words, i.e., $D =[ts_1, ..., ts_m]$ and $ts_i = [w_{i1}, ..., w_{in_i}]$, and a base model $\mathcal{M}$ with encoder-decoder architecture, the aim is to modify $\mathcal{M}$ to derive an extractive summary and an abstractive summary, i.e., $S_{ext}^*=[ts_i, ..., ts_j, ...., ts_k]$ where $1 \leq i < j < k \leq m$ and $S_{abs}^*=[w_1^{s^*}, ...., w_l^{s^*}]$. 

Additionally, let $\mathcal{S}$ denote the human-written reference summary for $D$ with $r$ words, i.e., $\mathcal{S} = [w_1^s, ..., w_r^s]$. The set of ground truth labels for each textual segment could be derived from $\mathcal{S}$, i.e., \(GT=[gt_1,\dots,gt_{m}]\), via the same greedy algorithm as applied in Section \ref{sec:mask}. If $ts_i$ is selected by the algorithm, $gt_i=1$; otherwise, $gt_i=0$.

\subsection{Model}
\label{section:model}
As illustrated in Figure \ref{Fig:model}, the proposed model $\textsc{ExtAbs}$ (pseudo-code in Appendix \ref{appendix:learning_algo}) has two modules. In the extractor, the entire input sequence goes through the encoder in $\mathcal{M}$ to get contextual representations for input tokens, and the self-attentive span extractor aggregates token representations to derive representations for each textual segment. Then, the classification layer predicts saliency scores for each textual segment. In the abstractor, a saliency mask is first generated for those textual segments with top-$z$ highest saliency scores. Then, the decoder generates summary tokens auto-regressively but attends to the input sequence based on the saliency mask. The extractive summary is formed with the textual segments with top-$k$ saliency scores, and the abstractive summary is the sequence generated by the decoder.


\paragraph{Extractor} Given a document, the encoder in $\mathcal{M}$ takes $D$ as input and outputs hidden states as contextual representations for each token in $D$:
\begin{gather}
    \resizebox{.85\hsize}{!}{$\{w_{11}^e, ..., w_{1n_1}^e, ..., w_{m1}^e, ..., w_{mn_m}^e\} = \mathcal{M}_{enc}(D)$} \label{equa:encoder}
\end{gather}
Following previous works that formulate the extractive task as sequence labelling, we introduce the span aggregation and classification layers. A self-attentive span extractor is applied to aggregate token representations for each textual segment based on the predicted token attentions, where a feed-forward network (FFN) calculates an attention score for each token and the softmax layer normalizes the scores, i.e.,
\begin{gather}
    \alpha_{ij} = \frac{exp(FFN(w_{ij}^e))}{\sum_{k=1}^{n_i}exp(FFN(w_{ik}^e))} \label{equa:span_attn} \\
    ts_i^e = \sum_{j=1}^{n_i}\alpha_{ij}w_{ij}^e \label{equa:span_agg}
\end{gather}
Lastly, the linear classification layer with the sigmoid function $\sigma$ and trainable parameters $W^c$ predicts a saliency score for each textual segment, i.e.,
\begin{gather}
    score_i = \sigma(W^c  ts_i^e) \label{equa:span_score}
\end{gather}

After ranking the predicted saliency scores, those textual segments with top-$k$ highest scores are concatenated in the order they appear in the document to form the extractive summary, i.e.,
\begin{gather}
    S_{ext}^* = [ts_{i_1}, ..., ts_{i_j}, ..., ts_{i_k}] \label{equa:span_ext}
\end{gather}
where \(i_j \leq m\) and \(score_{i_j} \in\) top-\({k}(\mathbf{score}), j=1,2,...,k\).

\paragraph{Abstractor} Following the convention of auto-regressive generation, the decoder in $\mathcal{M}$ generates a probability distribution $\mathbf{P}(w_t)$ over the pre-defined vocabulary at the $t$-th decoding step based on the encoder's hidden states of the input sequence and the previously decoded tokens. A token $w_t^{s^*}$ is sampled from the dictionary based on $\mathbf{P}(w_t)$ according to a specific decoding strategy. Differently, the vanilla encoder mask in the cross-attention module is replaced with a saliency mask, i.e.,
\begin{gather}
    \resizebox{\hsize}{!}{$\mathbf{P}(w_t|\mathbf{w}_{<t}^{s^*}, mask) = \mathcal{M}_{dec}([w_{11}^e, ..., w_{mn_m}^e], mask, \mathbf{w}_{<t}^{s^*})$} \label{equa:decoder}
\end{gather}
During training, $mask$ is determined like the preliminary analysis described in Section \ref{sec:mask}, i.e., the greedily selected salients based on reference summary. When conducting inference, $mask$ is determined by the predicted saliency scores, i.e., 
\begin{gather} \label{equa:topz_mask}
    mask_{ij} = \begin{cases}
        1, & \text{if $score_i \in$ top-$z(\mathbf{score})$} \\
         & \text{\phantom{if }and $j=1,2,...,n_i$} \\
        0, & \text{otherwise}
    \end{cases}
\end{gather}

\subsection{Loss Function}
A multi-task objective is adapted as the loss function $\mathcal{L}$ to train the model on both tasks jointly. The binary cross entropy between the predicted scores and ground truth labels is minimized for the extractive task, and the negative log-likelihood of the reference summary is minimized for the abstractive task. To stabilize the training, we additionally introduce the Kullback–Leibler (KL) divergence loss as a regularizer to prevent the large divergence caused by the saliency mask. The loss function is formulated as below.
\begin{gather}
    \resizebox{\hsize}{!}{$\mathcal{L}_{ext} =-\sum_{i=1}^{m}\Bigl(gt_i log(score_i) + (1-gt_i) log(1-score_i)\Bigr)$} \label{equa:ext_loss} \\
    \mathcal{L}_{abs} = -\sum_{i=1}^{r} logP(w_i^s|\mathbf{w}_{<i}^s, mask) \label{equa:abs_loss} \\
    \resizebox{\hsize}{!}{$\mathcal{L}_{KL} = \sum_{i=1}^{r} KL\Bigl(\mathbf{P}(w_i|\mathbf{w}_{<i}, mask) \parallel \mathbf{P}(w_i|\mathbf{w}_{<i})\Bigr)$} \label{equa:kl_loss} \\
    \mathcal{L} = \alpha \mathcal{L}_{ext} + \beta \mathcal{L}_{abs} + \gamma \mathcal{L}_{KL}
\end{gather}
where $\alpha, \beta$ and $\gamma$ are hyperparameters to balance the three terms.

\begin{table*}[ht!]
  \centering
  \resizebox{\textwidth}{!}
  {%
  \begin{tabular}{llcccccc}
    \hline
    \textbf{Dataset} & \textbf{Model} & \textbf{Top} & \textbf{ROUGE-1} & \textbf{ROUGE-2} & \textbf{ROUGE-L} & \textbf{BERTScore} & \textbf{BARTScore} \\
    \hline
    \multirow{14}{*}{\textbf{CNN/DM}} & Extractive SOTA$^{\dagger}$ & - & \textbf{44.80 }& \textbf{21.66} & \textbf{42.56} & - & - \\
    & $\textsc{MatchSum}^{\dagger}$ & - & 44.41 & 20.86 & 40.55 & - & - \\
    & $\textsc{Extractor}(\textsc{BART})$ & $k$=7 & 43.90 & 21.49 & 41.71 & 0.86 & -4.40 \\
    & $\textsc{Extractor}(\textsc{PEGASUS})$ & $k$=7 & 43.75 & 21.40 & 41.58 & 0.86 & -4.39 \\
    & $\textsc{ExtAbs}(\textsc{BART})$-ext & $k$=7 & 43.96 & 21.59 & 41.78 & 0.86 & -4.37 \\
    & $\textsc{ExtAbs}(\textsc{PEGASUS})$-ext & $k$=7 & 44.04 & 21.55 & 41.88 & 0.86 & -4.41 \\
    \cdashline{2-8}
    & $\textsc{BART}$ (ours) & - & 44.31 & 21.34 & 41.36 & 0.88 & -4.44 \\
    & $\textsc{PEGASUS}$ (ours) & - & 43.67 & 20.96 & 40.74 & 0.88 & -4.42 \\
    & $\textsc{GSum}^{\dagger}$ & - & \textbf{45.94} & \textbf{22.32} & \textbf{42.48} & - & - \\
    & $\textsc{GSum}$ (ours) & - & 45.69 & 22.28 & 42.38 & \textbf{0.89} & \textbf{-4.11} \\
    & $\textsc{EDURewriter}^{\dagger}$ & - & 43.09 & 20.24 & 40.52 & - & - \\
    & $\textsc{ContextRewriter}^{\dagger}$ & - & 43.52 & 20.57 & 40.56 & - & - \\
    & $\textsc{ExtAbs}(\textsc{BART})$-abs & $z$=8 & \underline{45.31} & 21.84 & \underline{42.28} & \underline{0.88} & \underline{-4.25} \\
    & $\textsc{ExtAbs}(\textsc{PEGASUS})$-abs & $z$=8 & 45.06 & \underline{22.02} & 42.09 & 0.88 & -4.35 \\ 
    \hline
    \multirow{11}{*}{\textbf{Reddit}} & Extractive SOTA$^{\dagger}$ & - & 27.01 & 7.06 & 22.70 & - & - \\
    & $\textsc{MatchSum}^{\dagger}$ & - & 25.09 & 6.17 & 20.13 & - & - \\
    & $\textsc{Extractor}(\textsc{BART})$ & $k$=5 & 27.94 & 7.63 & 23.59 & 0.84 & -4.96 \\
    & $\textsc{Extractor}(\textsc{PEGASUS})$ & $k$=5 & 28.05 & 7.50 & 23.56 & 0.84 & -4.98 \\
    & $\textsc{ExtAbs}(\textsc{BART})$-ext & $k$=5 & \textbf{28.51} & \textbf{8.10} & \textbf{23.99} & 0.84 & -4.95 \\
    & $\textsc{ExtAbs}(\textsc{PEGASUS})$-ext & $k$=5 & 28.00 & 7.73 & 23.52 & 0.84 & -4.98 \\
    \cdashline{2-8}
    & $\textsc{BART}$ (ours) & - & 33.01 & \underline{11.63} & \underline{26.93} & 0.88 & \underline{-4.70} \\
    & $\textsc{PEGASUS}$ (ours) & - & 31.52 & 11.13 & 25.83 & 0.88 & -4.77 \\
    & $\textsc{GSum}^{\dagger}$ & - & \textbf{34.52} & \textbf{12.71} & \textbf{27.58} & - & - \\
    & $\textsc{ExtAbs}(\textsc{BART})$-abs & $z$=8 & \underline{33.42} & 11.41 & 26.68 & \textbf{0.88} & \textbf{-4.49} \\
    & $\textsc{ExtAbs}(\textsc{PEGASUS})$-abs & $z$=8 & 31.84 & 10.16 & 25.41 & 0.88 & -4.73 \\ 
    \hline
    \multirow{10}{*}{\textbf{PubMed}} & Extractive SOTA$^{\dagger}$ & - & 43.08 & 16.71 & 38.30 & - & -  \\
    & $\textsc{MatchSum}^{\dagger}$ & - & 41.21 & 14.91 & 36.75 & - & - \\
    & $\textsc{Extractor}(\textsc{BART})$ & $k$=22 & 43.48 & 17.27 & 40.72 & 0.85 & -4.56 \\
    & $\textsc{Extractor}(\textsc{PEGASUS})$ & $k$=22 & 43.54 & 17.22 & 40.78 & 0.85 & -4.56 \\
    & $\textsc{ExtAbs}(\textsc{BART})$-ext & $k$=22 & 43.48 & 17.29 & 40.73 & 0.85 & -4.56 \\
    & $\textsc{ExtAbs}(\textsc{PEGASUS})$-ext & $k$=22 & \textbf{43.71} & \textbf{17.42} & \textbf{40.93} & 0.85 & -4.55 \\
    \cdashline{2-8}
    & $\textsc{BART}$ (ours) & - & 43.57 & 16.47 & 40.17 & 0.86 & -4.63 \\
    & $\textsc{PEGASUS}$ (ours) & - & 43.41 & \underline{17.13} & 39.93 & 0.86 & -4.64 \\
    & $\textsc{GSum}^{\dagger}$ & - & \textbf{45.09} & 16.72 & \textbf{41.32} & - & -  \\
    & $\textsc{ExtAbs}(\textsc{BART})$-abs & $z$=25 & \underline{43.90} & 16.12 & \underline{40.49} & \textbf{0.86} & \underline{-4.41} \\
    & $\textsc{ExtAbs}(\textsc{PEGASUS})$-abs & $z$=25 & 43.61 & \textbf{17.26} & 40.16 & 0.86 & \textbf{-4.39} \\ 
    \hline
  \end{tabular} %
  }
  \caption{Experimental results on test sets of three datasets. $\textsc{ExtAbs}$(*) refers to our adapted version of the corresponding vanilla encoder-decoder model *. ext and abs refer to the extractive and abstractive results, respectively. $^{\dagger}$ indicates that the results are copied from the corresponding original paper. $k$ and $z$ refer to the number of extracted textual segments for extractive summary and saliency mask, respectively, determined by validation sets. The \textbf{best} and \underline{second} scores within each block are bold and underlined, respectively.}
  \label{Tab:experiments}
\end{table*}

\section{Experiments}

\subsection{Experimental Setup}

\paragraph{Datasets} \textbf{CNN/DailyMail} (CNN/DM) \cite{10.5555/2969239.2969428} is a widely used dataset for summarization tasks. Each news article comes with several highlight sentences written by humans as the reference summary in the dataset. \textbf{Reddit} \cite{kim-etal-2019-abstractive} is crawled from the social media forum Reddit where the content in the crawled post and TL;DR are treated as the document and reference summary, respectively. \textbf{PubMed} \cite{cohan-etal-2018-discourse} is collected from the scientific paper repository PubMed.com, where the abstract is taken as the reference summary. Following \citet{zhong-etal-2020-extractive}, we use the truncated version of PubMed with the introduction section in the paper as the document. The detailed statistics about the three datasets are listed in Appendix \ref{appendix:dataset}.

\paragraph{Baselines} \textbf{$\textsc{BART}$} \cite{lewis-etal-2020-bart} and \textbf{$\textsc{PEGASUS}$} \cite{10.5555/3524938.3525989} are the two most widely adopted pre-trained encoder-decoder models for abstractive summarization, and our proposed $\textsc{ExtAbs}$ is built upon them. We also include the \textbf{$\textsc{Extractor}$} in $\textsc{ExtAbs}$ built upon their encoders only as extractive baselines. \textbf{$\textsc{ContextRewriter}$} \cite{bao2021contextualized} and \textbf{$\textsc{EDURewriter}$} \cite{10.1145/3477495.3531916} are two extract-then-abstract models with the same highlight method but the former one highlights at sentence level, while the latter one highlights EDUs. \textbf{$\textsc{GSum}$} \cite{dou-etal-2021-gsum} extends $\textsc{BART}$ with an extra encoder for highlighting purposes and achieves superior performance on multiple datasets. Notably, $\textsc{GSum}$ serves as the abstractor in the extract-then-abstract paradigm only when it takes the extracted sentences as guidance. \textbf{$\textsc{MatchSum}$} \cite{zhong-etal-2020-extractive} is the extractor of salient sentences used in $\textsc{GSum}$. \textbf{$\textsc{EDU-VL}$} \cite{wu-etal-2023-edu} is an extractive model that extracts EDUs from the input document and achieves SOTA results on CNN/DM and Reddit. \textbf{$\textsc{MemSum}$} \cite{gu-etal-2022-memsum} is a reinforcement learning-based extractive model, achieving SOTA performance on PubMed. We also include \textbf{$\textsc{GPT-4}$} \cite{achiam2023gpt} as a strong LLM baseline by following \citet{zhang-etal-2023-extractive-summarization}'s work to prompt it to perform extractive, abstractive and extract-then-abstract summarization tasks on a randomly selected subset for each dataset.

\paragraph{Evaluation Metrics} Automatic and human evaluations are conducted to evaluate the model performance comprehensively. The automatic evaluation metrics include ROUGE-1/2/L \cite{lin-2004-rouge}, BERTScore \footnote{ROUGE scores and BERTScore are calculated by the HF library \url{https://huggingface.co/evaluate-metric}. In this work, the ROUGE-LSum score is reported to align with previous works. F1 score is reported.}\cite{DBLP:conf/iclr/ZhangKWWA20} and BARTScore \footnote{We use the trained ParaBank version BARTScore.}\cite{DBLP:conf/nips/YuanNL21} to measure from lexical and semantic perspectives.
Human evaluation metrics include factuality, informativeness and ranking.

\paragraph{Implementation Details} All models are trained using Pytorch on up to four A100 80G GPUs. The checkpoint with the best ROUGE-L score on the validation set is taken as the final model for each experiment. More implementation details about hyperparameters are provided in Appendix \ref{appendix:imp_details}.

\subsection{Results}

Main experimental results are presented in Table \ref{Tab:experiments}, and the corresponding statistical significance test to determine if an improvement is significant is presented in Appendix \ref{appendix:sig_test}. The BERTScore for all models are very close to each other; therefore, we will ignore them from the discussion.  

\paragraph{Main Results} On CNN/DM, while the proposed $\textsc{ExtAbs}(*)$ underperforms the SOTA extractive model on all ROUGE scores, it significantly outperforms $\textsc{MatchSum}$ on ROUGE-2/L. In the abstractive task, $\textsc{ExtAbs}(\textsc{BART})$ and $\textsc{ExtAbs}(\textsc{PEGASUS})$ significantly outperform their vanilla counterparts on all ROUGE scores. $\textsc{ExtAbs}(*)$ also surpasses other extract-then-abstract models like $\textsc{EDURewriter}$ and $\textsc{ContextRewriter}$. Though $\textsc{GSum}$ achieves the best results, it is noteworthy that $\textsc{GSum}$ primarily focuses on abstractive performance, whereas $\textsc{ExtAbs}$ seamlessly unifies extractive and abstractive summarization within a single model.
On Reddit and PubMed, $\textsc{ExtAbs}(*)$ outperforms the SOTA extractive model and significantly outperforms $\textsc{MatchSum}$ on all ROUGE scores, while achieving comparable scores to the vanilla abstractive baseline, i.e., the scores on some metrics are higher while some are lower.
Besides, varying degrees of improvement are observed when comparing $\textsc{ExtAbs}(*)$ to the corresponding $\textsc{Extractor}(*)$ across three datasets, indicating that joint training enhances extractive performance.

\begin{table}[t]
  \centering
  \small
  \resizebox{\columnwidth}{!}
  {%
  \begin{tabular}{llcccc}
    \toprule
    \textbf{Model} & \textbf{Task} & \textbf{R-1} & \textbf{R-2} & \textbf{R-L} & \textbf{BS} \\
    \hline
    \multicolumn{6}{c}{\textbf{CNN/DailyMail}} \\
    \hline
    \multirow{2}{*}{$\textsc{ExtAbs}$}& ext & 42.99 & 19.70 & 40.67 & -4.43 \\
     & abs & \textbf{44.26} & \textbf{21.31} & \textbf{41.82} & \textbf{-4.31} \\
    \cdashline{1-6}
    \multirow{3}{*}{$\textsc{GPT-4}$} & ext & \underline{38.60} & \underline{14.65} & \underline{32.07} & -4.89 \\
    & abs & 36.17 & 12.68 & 28.45 & \underline{-4.80} \\
    & ext-abs & 35.13 & 12.16 & 27.73 & -4.92 \\
    \hline
    \multicolumn{6}{c}{\textbf{Reddit}} \\
    \hline
    \multirow{2}{*}{$\textsc{ExtAbs}$}& ext & 27.80 & 9.08 & 23.87 & -4.86 \\
     & abs & \textbf{34.40} & \textbf{12.31} & \textbf{26.86} & \textbf{-4.64} \\
    \cdashline{1-6}
    \multirow{3}{*}{$\textsc{GPT-4}$} & ext & \underline{26.94} & 6.83 & \underline{19.40} & \underline{-4.95} \\
    & abs & 25.97 & \underline{6.97} & 18.63 & -4.95 \\
    & ext-abs & 24.01 & 5.18 & 16.99 & -5.15 \\
    \hline
    \multicolumn{6}{c}{\textbf{PubMed}} \\
    \hline
    \multirow{2}{*}{$\textsc{ExtAbs}$}& ext & \textbf{44.79} & \textbf{18.47} & \textbf{42.09} & \textbf{-4.58} \\
     & abs & 44.72 & 17.33 & 40.98 & -4.67 \\
    \cdashline{1-6}
    \multirow{3}{*}{$\textsc{GPT-4}$} & ext & \underline{40.86} & \underline{14.62} & \underline{32.19} & \underline{-5.11} \\
    & abs & 37.30 & 11.47 & 29.28 & -5.25 \\
    & ext-abs & 40.16 & 12.62 & 31.06 & -5.14 \\
    \bottomrule
  \end{tabular} %
  }
  \caption{Results on 50 randomly sampled instances from test sets. Here BS refers to BARTScore.}
  \label{Tab:gpt_results}
\end{table}

\paragraph{GPT-4 Results} As shown in Table \ref{Tab:gpt_results}, $\textsc{GPT-4}$ underperforms $\textsc{ExtAbs}(\textsc{BART})$ on all extractive-only, abstractive-only and extract-then-abstract settings on all metrics.


\paragraph{Discussion} We attribute the inconsistent abstractive performance of $\textsc{ExtAbs}$ across the three datasets to the quality of extractive summaries from the extractor and the potential boost from the proposed saliency mask. The average ROUGE score gain is significantly higher on CNN/DM (13.35) than on Reddit (6.22) and PubMed (2.24) when inferring with saliency mask derived from the reference summary, suggesting a larger potential boost on CNN/DM. The comparable abstractive performance of $\textsc{ExtAbs}$ on Reddit is likely due to lower extractive summary quality, while PubMed shows less potential boost from the saliency mask.
More details are provided in Appendix \ref{appendix:discussion}.

\subsection{Ablation Analysis}

Results of ablation analysis are shown in Table \ref{Tab:ablation}.

\begin{table}[t]
  \centering
  \resizebox{\columnwidth}{!}
  {%
  \begin{tabular}{llcccc}
    \hline
    \textbf{Hyperparameter} & \textbf{Task} & \textbf{R-1} & \textbf{R-2} & \textbf{R-L} & \textbf{BS} \\
    \hline
    \multirow{2}{*}{$\textsc{ExtAbs}(\textsc{BART})$} & ext & 43.96 & 21.59 & 41.78 & -4.37 \\
    & abs & \textbf{45.31} & \textbf{21.84} & \textbf{42.28} & \textbf{-4.25} \\
    \cdashline{2-6}
    \multirow{2}{*}{input = sentences} & ext & 43.47 & 20.82 & 39.95 & -4.44 \\
    & abs & 45.14 & 21.80 & 42.08 & -4.26 \\
    \cdashline{2-6}
    \multirow{2}{*}{$\alpha=50$} & ext & 44.07 &  21.62 & 41.86 & -4.38 \\
    & abs & 45.01 & 21.60 & 41.98 & -4.28 \\
    \cdashline{2-6}
    \multirow{2}{*}{w/o $mask$} & ext & 42.82 &  20.51 & 40.70 & -4.41 \\
    & abs & 44.78 & 21.59 & 41.72 & -4.26 \\
    \hline
  \end{tabular} %
  }
  \caption{Results of the ablation analysis on the proposed $\textsc{ExtAbs}(\textsc{BART})$ with different input granularities and $\alpha$ values, with and without saliency mask.}
  \label{Tab:ablation}
\end{table}


\paragraph{Granularity of Highlighted Information} The sentence-level $\textsc{ExtAbs}(\textsc{BART})$, using sentences as textual segments, is trained to compare with the EDU-level one. Abstractive summaries generated by the sentence-level model achieve scores comparable to those of the EDU-level model. However, there is a significant decrease in all ROUGE scores and BARTScore for extractive summaries derived from the sentence-level model. This observation echoes the conclusion drawn by \citet{li-etal-2016-role} and \citet{wu-etal-2023-edu}, i.e., EDU is a better text unit for extractive summarization.

\paragraph{Extractive vs. Abstractive} We further tune the hyperparameter $\alpha$ to validate the influence made by the loss function. The increase of $\alpha$ (weight of extractive loss) results in better scores for the extractive summary while lower scores for the abstractive summary. The result suggests a tradeoff between extractive and abstractive summarization performances within one single model.

\paragraph{Saliency Mask} To validate the effectiveness of the proposed saliency mask in $\textsc{ExtAbs}$, we compare model performance with and without the proposed saliency mask. It is observed that there is a decrease in all four metrics for both extractive and abstractive summaries, demonstrating the necessity of the saliency mask.

\subsection{Human Evaluation}

\begin{table}[]
    \small
    \centering
    \begin{tabular}{lccc}
    \hline
        \textbf{Model} & \textbf{Fact.} & \textbf{Info.} & \textbf{Ranking} \\
        \hline
        $\textsc{BART}$ & 0.80 & 0.20 & 2.20 \\
        $\textsc{ExtAbs}(\textsc{BART})$-ext & \textbf{1.00} & \textbf{0.37} & 2.07 \\
        $\textsc{ExtAbs}(\textsc{BART})$-abs & 0.70 & \textbf{0.37} & \textbf{1.73} \\
    \hline
    \end{tabular}
    \caption{Human evaluation results on sampled instances.}
    \label{tab:human_eval}
\end{table}

 We randomly sample 30 CNN/DM test instances to compare summaries generated by the baseline $\textsc{BART}$ and our proposed $\textsc{ExtAbs}(\textsc{BART})$. Annotators are asked to rate either 0 or 1 to indicate whether the generated summary is faithful to the source document (factuality) and contains all salient information from the reference summary (informativeness), and rank among the three summaries for the overall quality of a summary. Table \ref{tab:human_eval} presents the averaged human evaluation results. 
 Firstly, the extractive summaries are entirely factual, while varying degrees of hallucination are observed in abstractive summaries, aligning with expectations for extractive summaries. Secondly, informativeness scores are relatively low across all summary types, indicating the challenge of comprehensively capturing all salient information. Overall, abstractive summaries generated by our proposed $\textsc{ExtAbs}(\textsc{BART})$ achieve the lowest ranking value, suggesting annotators' preference over the baseline. Additional evaluation results from $\textsc{GPT-4}$ of these samples on more aspects are in Appendix \ref{Appendix:gpt_evaluation}.

\section{Conclusion}
In this paper, we discover and highlight the importance of highlighting salient information in the extract-then-abstract paradigm by applying the saliency mask in the decoder of the abstractor. Our proposed saliency mask is parameter-free and achieves higher ROUGE scores than other highlight methods. Then, we propose $\textsc{ExtAbs}$, an extract-and-abstract framework to unify any encoder-decoder model to jointly and seamlessly perform extractive and abstractive summarization tasks. In $\textsc{ExtAbs}$, the encoder is augmented and serves as the extractor, and the decoder along with the encoder serves as the abstractor. Our experiments on Reddit and PubMed demonstrate that the proposed method generates better extractive summaries and performs comparable, or even better than the vanilla model on the abstractive task.

\section{Limitations}
Firstly, the proposed $\textsc{ExtAbs}$ has only been tested on $\textsc{BART}$ and $\textsc{PEGASUS}$, but we acknowledge that there are other widely used pre-trained encoder-decoder models, such as the $\textsc{T5}$ family. It could be worthwhile to conduct experiments with more baseline models given sufficient time and resources. Secondly, the proposed highlight method, i.e., saliency mask, can only be applied to the encoder-decoder models and cannot be extended to the decoder-only models directly, e.g., the $\textsc{GPT}$ family. Considering the recent popularity of the decoder-only model, it is worth exploring a compatible way for the decoder-only models, such as integrating the saliency mask with the original self-attention mask. We leave such exploration for future work.



\bibliography{custom}

\appendix

\section{Greedy Selection Algorithm}
\label{appendix: greedy algo}
Algorithm \ref{algo: greedy} presents the pseudo-code of the algorithm of selecting salient textual segments, which is used to generate saliency masks and ground truth labels.

\begin{algorithm}
\caption{Greedy Selection Algorithm}
\label{algo: greedy}
\begin{algorithmic}[1]
\Require{$Doc, Ref, k$} \Comment{$k$: \# of selections}
\Ensure{$sel\_idx$} \Comment{selected indices}
\State {$sel\_idx$ $\gets$ [ ]} \Comment{empty list}
\State {$C \gets $ [ ]} \Comment{candidate: empty list}
\While{$k \geq 0$}
\State {end $\gets$ TRUE }
\For{$i \gets 0$ to $len(Doc)$}
  \State {$tmp\_C$ $\gets$ $ C + [Doc_i]$}
  \State {$score$ $\gets$ $ROUGE(tmp\_C, Ref)$}
  \If {$score$ increases}
    \State {$sel\_idx$ $\gets$ $sel\_idx + [i]$}
    \State {$C \gets tmp\_C$}
    \State {$k \gets k-1$}
    \State {end $\gets$ FALSE}
    \State break
\EndIf
\EndFor
\If {end}
  \State break
\EndIf
\EndWhile
\State \textbf{return} $sel\_idx$
\end{algorithmic}
\end{algorithm}

\section{Learning Algorithm}
\label{appendix:learning_algo}
Algorithm \ref{algo:extabs_learning} summarizes the model learning procedure in alignment with the description in Section \ref{section:model}. 

\begin{algorithm*}
  \caption{\textbf{Model Learning Algorithm}}
  \label{algo:extabs_learning}
  \begin{algorithmic}[1]
  \Require{$D, \mathcal{M}, k, z, GT, S$} \Comment{\small $D=[w_{11}, ..., w_{1n_1}, ..., w_{m1}, ..., w_{mn_m}]=[edu_1,...,edu_m]; GT=[gt_1,..., gt_m]$}
  \Ensure{$S^*_{ext}, S^*_{abs}$}
  \State tokenRep$\vert_{11}^{mn_m}$ $\gets \mathcal{M}_{enc}(D)$ \Comment{Equation (\ref{equa:encoder})}
  \For{$i \gets 1$ to $m$}
    \State eduRep$_i \gets$ \textit{SpanExtractor}(tokenRep$\vert_{i1}^{in_i}$) \Comment{Equation (\ref{equa:span_attn}) and Equation (\ref{equa:span_agg})}
  \EndFor
  \State score$\vert_1^m \gets $ \textit{ClassificationLayer}(eduRep$\vert_1^m$) \Comment{Equation (\ref{equa:span_score})}
  \State $S^*_{ext} \gets []$
  \For{$i \in$ indices of \textit{Top}-$k$(score$\vert_1^m$)} 
    \State $S^*_{ext} \gets S^*_{ext} + $ [edu$_{i}$] \Comment{Equation (\ref{equa:span_ext}): form extractive summary}
  \EndFor

  \State mask$\vert_{11}^{mn_m} \gets 0$ \Comment{Initialise saliency mask}
  \If{training}
    \For {$j \gets 1$ to $m$} \Comment{Equation (\ref{equa:oracle_mask})}
      \If{$gt_j = 1$} 
        \State mask$_{j*} \gets 1$ 
      \EndIf
    \EndFor
  \Else
    \For {$j \in$ indices of \textit{Top}-$z$(score$\vert_1^m$)} \Comment{Equation (\ref{equa:topz_mask})}
      \State mask$_{j*} \gets 1$ 
    \EndFor
  \EndIf
  \State $P\vert_1^r$ or $P\vert_1^l \gets \mathcal{M}_{dec}$(tokenRep, mask) \Comment{Equation (\ref{equa:decoder})}
  \State $S^*_{abs} \sim P\vert_1^l$ \Comment{Derive abstractive summary}

  \If{training}
    \State $\mathcal{L}_{ext} \gets$ binary cross entropy between score$\vert_1^m$ and $GT\vert_1^m$ \Comment{Equation (\ref{equa:ext_loss})}
    \State $\mathcal{L}_{abs} \gets$ negative log-likelihood of $S \sim P\vert_1^r$ \Comment{Equation (\ref{equa:abs_loss})}
    \State $P'\vert_1^r \gets \mathcal{M}_{dec}$(tokenRep)
    \State $\mathcal{L}_{KL} \gets$ KL divergence between $P\vert_1^r$ and $P'\vert_1^r$ \Comment{Equation (\ref{equa:kl_loss})}
    \State Update parameters based on $\mathcal{L}_{ext}, \mathcal{L}_{abs}$ and $\mathcal{L}_{KL}$
  \EndIf

  \State \textbf{return} $S^*_{ext}, S^*_{abs}$
  \end{algorithmic}
\end{algorithm*}

\section{Dataset Statistics}
\label{appendix:dataset}
The statistics of each dataset are listed in Table \ref{Tab:datasets}.
\begin{table}
  \centering
  \resizebox{\columnwidth}{!}
  {%
  \begin{tabular}{lcccccc}
    \hline
    \multirow{2}{*}{Dataset} & \multicolumn{3}{c}{\#Pairs} & & \multicolumn{2}{c}{\#Tokens} \\
    \cmidrule{2-4}\cmidrule{6-7}
    & Train & Valid & Test & & Doc. & Sum. \\
    \hline
    CNN/DM & 287,226 & 13,368 & 11,490 & & 766 & 58 \\
    Reddit & 41,694 & 645 & 645 & & 482 & 28 \\
    PubMed & 87,445 & 4,928 & 4,986 & & 444 & 210 \\
    \hline
  \end{tabular} %
  }
  \caption{Dataset statistics.}
  \label{Tab:datasets}
\end{table}

\section{Implementation Details}
\label{appendix:imp_details}
We follow \citet{xu-etal-2020-discourse} and \citet{wu-etal-2023-edu} to do the EDU segmentation.

Regarding the preliminary experiments in Table \ref{Tab:preliminary}, we follow the setup in \citet{liu-lapata-2019-text} except for the adjustment of batch size to 2800 to fit the GPU memory best, and we set the maximum number of oracle sentences and EDUs for our saliency mask as 5 and 8, respectively. 

Regarding the experiments for $\textsc{ExtAbs}$ in Table \ref{Tab:experiments}, the textual segment is EDU. For the experiments about $\textsc{PEGASUS}$-based models, we fine-tune or adapt the "pegasus-large" model and follow the same learning rate, length penalty, number of beams, etc., for each dataset as reported by \citet{10.5555/3524938.3525989}. For all $\textsc{BART}$-based models, we fine-tune or adapt the "bart-large" model. The learning rate and the number of beams are set to be 1e-5 and 4 on all datasets, respectively, while batch size varies. The values for $\alpha$ and $\beta$ also vary between datasets and base models. The default value for $\gamma$ is 0.0 to ignore the KL divergence loss, and the only exception is $\textsc{ExtAbs}(\textsc{BART})$ for Reddit where $\gamma$ is set to 0.01. Such an exception is determined by the checkpoint's performance on the validation set. For example, on the CNN/DM dataset, though the results of the model with KL divergence and without KL divergence are quite close (ROUGE-1: 45.18 vs 45.31; ROUGE-2: 21.69 vs 21.84; ROUGE-L: 42.23 vs 42.28), only the best one on the validation set would be reported. Depending on the size of the training dataset and the number of GPUs used for training, the running time for each experiment varies between 1 to 4 days. The specific hyperparameter values for each experiment are in Table \ref{Tab:exp_details}. 

\begin{table*}[ht!]
  \centering
  \resizebox{\textwidth}{!}
  {%
  \begin{tabular}{llccccccccc}
    \hline
    \multirow{2}{*}{\textbf{Dataset}} & \multirow{2}{*}{\textbf{Model}} & \textbf{Max input} & \textbf{Max target} & \textbf{Learning} & \textbf{Batch} & \textbf{Beam} & \textbf{Length} & \multirow{2}{*}{$\mathbf{\alpha}$} & \multirow{2}{*}{$\mathbf{\beta}$} & \multirow{2}{*}{$\mathbf{\gamma}$} \\
     &  & \textbf{token} & \textbf{token} & \textbf{rate} & \textbf{size} & \textbf{size} & \textbf{penalty} &  &  &   \\
    \hline
    \multirow{6}{*}{\textbf{CNN/DM}} & $\textsc{Extractor}(\textsc{BART})$ & 1024 & - & 1e-5 & 16 & - & - & - & - & - \\
    & $\textsc{BART}$ & 1024 & 128 & 1e-5 & 16 & 4 & 1.0 & - & - & - \\
    & $\textsc{ExtAbs}(\textsc{BART})$ & 1024 & 128 & 1e-5 & 16 & 4 & 1.0 & 10.0 & 1.0 & 0.0 \\
    & $\textsc{Extractor}(\textsc{PEGASUS})$ & 1024 & - & 5e-5 &  & - & - & - & - & - \\
    & $\textsc{PEGASUS}$ & 1024 & 128 & 5e-5 & 8 & 8 & 0.9 & - & - & - \\
    & $\textsc{ExtAbs}(\textsc{PEGASUS})$ & 1024 & 128 & 5e-5 & 8 & 8 & 0.9 & 10.0 & 1.0 & 0.0  \\
    \cdashline{2-11}
    \multirow{6}{*}{\textbf{Reddit}} & $\textsc{Extractor}(\textsc{BART})$ & 1024 & - & 1e-5 & 16 & - & - & - & - & - \\
    & $\textsc{BART}$ & 1024 & 32 & 1e-5 & 16 & 4 & 1.0 & - & - & - \\
    & $\textsc{ExtAbs}(\textsc{BART})$ & 1024 & 32 & 1e-5 & 16 & 4 & 1.0 & 10.0 & 0.5 & 0.01  \\
    & $\textsc{Extractor}(\textsc{PEGASUS})$ & 1024 & - & 1e-4 & 16 & - & - & - & - & - \\
    & $\textsc{PEGASUS}$ & 1024 & 64 & 1e-4 & 8 & 8 & 0.6 & - & - & - \\
    & $\textsc{ExtAbs}(\textsc{PEGASUS})$ & 1024 & 64 & 1e-4 & 8 & 8 & 0.6 & 1.0 & 1.0 & 0.0  \\
    \cdashline{2-11}
    \multirow{6}{*}{\textbf{PubMed}} & $\textsc{Extractor}(\textsc{BART})$ & 1024 & - & 1e-5 & 16 & - & - & - & - & - \\
    & $\textsc{BART}$ & 1024 & 256 & 1e-5 & 16 & 4 & 1.0 & - & - & - \\
    & $\textsc{ExtAbs}(\textsc{BART})$ & 1024 & 256 & 1e-5 & 8 & 4 & 1.0 & 1.0 & 1.0 & 0.0 \\
    & $\textsc{Extractor}(\textsc{PEGASUS})$ & 1024 & - & 2e-4 & 16 & - & - & - & - & - \\
    & $\textsc{PEGASUS}$ & 1024 & 256 & 2e-4 & 8 & 8 & 0.8 & - & - & -  \\
    & $\textsc{ExtAbs}(\textsc{PEGASUS})$ & 1024 & 256 & 2e-4 & 8 & 8 & 0.8 & 1.0 & 10.0 & 0.0  \\
    \hline
  \end{tabular} %
  }
  \caption{Implementation details for each experiment in Table \ref{Tab:experiments}.}
  \label{Tab:exp_details}
\end{table*}

For experiments on $\textsc{GPT-4}$ in Table \ref{Tab:gpt_results}, we randomly sample 50 test instances for each dataset and adapt the prompts designed by \citet{zhang-etal-2023-extractive-summarization}. Three examples from the corresponding training set are selected for the few-shot learning of the extractive task. 
We experiment on the "gpt-4-turbo"\footnote{\url{https://platform.openai.com/docs/models/gpt-4-turbo-and-gpt-4}} model and set the temperature as 0 to ensure reproductivity. The prompts for each task are provided in Table \ref{Tab:appendix_prompts}.

\begin{table*}[ht!]
    \centering
    \begin{tabular}{p{0.18\textwidth}p{0.78\textwidth}}
        \hline
        \textbf{\small Task} & \textbf{\small Prompt} \\
        \hline
         \multirow{2}{*}{{\small Extractive}} & {\small \textbf{System:} You are an extractive summarizer that follows the output pattern.} \\
          & {\small \textbf{User:} The following examples are successful extractive summarization instances: \textit{<3-shot document-oracle summary pairs>}. Please summarize the following document. Document: \textit{<document>}. The summary should contain $m$ sentences. Provide the summary below: } \\
         \hline
         \multirow{2}{*}{{\small Abstractive}} & {\small \textbf{System:} You are an abstractive summarizer that follows the output pattern.} \\
          & {\small \textbf{User:} Please write a summary for the document. Document: \textit{<document>}. Provide the summary below: } \\
         \hline
         {\small Abstractive} & {\small \textbf{System:} You are an abstractive summarizer that follows the output pattern.} \\
         {\small (Reddit)} & {\small \textbf{User:} Please write a TL;DR for the Reddit post. Post: \textit{<document>}. Provide the TL;DR below: } \\
         \hline
         \multirow{2}{*}{{\small Extract-then-Abstract}} & {\small \textbf{System:} You are an abstractive summarizer that follows the output pattern.} \\
          & {\small \textbf{User:} Please revise the extracted summary based on the document. The revised summary should include the information in the extracted summary. Document: \textit{<document>}. Extracted Summary: \textit{<GPT-4-generated extractive summary>}. Provide the revised summary below: } \\
         \hline
    \end{tabular}
    \caption{Prompts for $\textsc{GPT-4}$.}
    \label{Tab:appendix_prompts}
\end{table*}



\section{Significance Testing}
\label{appendix:sig_test}
We conduct the t-test at a significance level of 0.05 to determine if model \#1 achieves significantly higher metric scores than model \#2. For extractive summaries, we compare $\textsc{ExtAbs}(*)$ with $\textsc{MatchSum}$ as $\textsc{MatchSum}$ released their extractive summaries. For abstractive summaries, we compare $\textsc{ExtAbs}(\textsc{*})$ with the corresponding vanilla model $*$. The results for the three datasets are presented in Table \ref{Tab:appendix_significance}.  

\begin{table*}[ht!]
    \centering
    \begin{tabular}{lcccccc}
        \hline
        \textbf{\small Dataset} & \textbf{\small Task} & \textbf{\small{Model \#1}} & \textbf{\small{Model \#2}} & \textbf{\small Metric} & \textbf{\small t-test score} & \textbf{\small p-value} \\
        \hline
        \multirow{12}{*}{\small \textbf{CNN/DM}} & \multirow{6}{*}{\small Extractive} & \small{$\textsc{MatchSum}$} & \small{$\textsc{ExtAbs}(\textsc{BART})$} & \small ROUGE-1 F1 & \small 2.29 & \small \textbf{0.01} \\
         & & \small{$\textsc{ExtAbs}(\textsc{BART})$} & \small{$\textsc{MatchSum}$} & \small ROUGE-2 F1 & \small 11.13 & \small \textbf{6.51e-29} \\  
         & & \small{$\textsc{ExtAbs}(\textsc{BART})$} & \small{$\textsc{MatchSum}$} & \small ROUGE-L F1 & \small 18.02 & \small \textbf{8.18e-72} \\ 
         & & \small{$\textsc{MatchSum}$} & \small{$\textsc{ExtAbs}(\textsc{PEGASUS})$} & \small ROUGE-1 F1 & \small 1.25 & \small 0.10 \\ 
         & & \small{$\textsc{ExtAbs}(\textsc{PEGASUS})$} & \small{$\textsc{MatchSum}$} & \small ROUGE-2 F1 & \small 9.86 & \small \textbf{3.76e-23} \\
         & & \small{$\textsc{ExtAbs}(\textsc{PEGASUS})$} & \small{$\textsc{MatchSum}$} & \small ROUGE-L F1 & \small 18.33 & \small \textbf{3.19e-74} \\ 
        \cdashline{2-7}
         & \multirow{6}{*}{\small Abstractive} & \small{$\textsc{ExtAbs}(\textsc{BART})$} & \small{$\textsc{BART}$} & \small ROUGE-1 F1 & \small 11.13 & \small \textbf{6.27e-29} \\ 
         & & \small{$\textsc{ExtAbs}(\textsc{BART})$} & \small{$\textsc{BART}$} & \small ROUGE-2 F1 & \small 5.69 & \small \textbf{6.44e-09} \\  
         & & \small{$\textsc{ExtAbs}(\textsc{BART})$} & \small{$\textsc{BART}$} & \small ROUGE-L F1 & \small 10.49 & \small \textbf{6.57e-26} \\ 
         & & \small{$\textsc{ExtAbs}(\textsc{PEGASUS})$} & \small{$\textsc{PEGASUS}$} & \small ROUGE-1 F1 & \small 14.33 & \small \textbf{1.72e-46} \\ 
         & & \small{$\textsc{ExtAbs}(\textsc{PEGASUS})$} & \small{$\textsc{PEGASUS}$} & \small ROUGE-2 F1 & \small 11.16 & \small \textbf{4.75e-29} \\  
         & & \small{$\textsc{ExtAbs}(\textsc{PEGASUS})$} & \small{$\textsc{PEGASUS}$} & \small ROUGE-L F1 & \small 13.83 & \small \textbf{1.90e-43} \\ 
        \hline
        \multirow{12}{*}{\small \textbf{Reddit}} & \multirow{6}{*}{\small Extractive} & \small{$\textsc{ExtAbs}(\textsc{BART})$} & \small{$\textsc{MatchSum}$} & \small ROUGE-1 F1 & \small 7.82 & \small \textbf{1.15e-14} \\
         & & \small{$\textsc{ExtAbs}(\textsc{BART})$} & \small{$\textsc{MatchSum}$} & \small ROUGE-2 F1 & \small 8.48 & \small \textbf{8.25e-17} \\  
         & & \small{$\textsc{ExtAbs}(\textsc{BART})$} & \small{$\textsc{MatchSum}$} & \small ROUGE-L F1 & \small 13.43 & \small \textbf{1.51e-36} \\ 
         & & \small{$\textsc{ExtAbs}(\textsc{PEGASUS})$} & \small{$\textsc{MatchSum}$} & \small ROUGE-1 F1 & \small 5.98 & \small \textbf{1.88e-09} \\ 
         & & \small{$\textsc{ExtAbs}(\textsc{PEGASUS})$} & \small{$\textsc{MatchSum}$} & \small ROUGE-2 F1 & \small 6.09 & \small \textbf{1.02e-09} \\
         & & \small{$\textsc{ExtAbs}(\textsc{PEGASUS})$} & \small{$\textsc{MatchSum}$} & \small ROUGE-L F1 & \small 11.10 & \small \textbf{1.63e-26} \\ 
        \cdashline{2-7}
         & \multirow{6}{*}{\small Abstractive} & \small{$\textsc{ExtAbs}(\textsc{BART})$} & \small{$\textsc{BART}$} & \small ROUGE-1 F1 & \small 1.03 & \small 0.15 \\ 
         & & \small{$\textsc{BART}$} & \small{$\textsc{ExtAbs}(\textsc{BART})$} & \small ROUGE-2 F1 & \small 0.53 & \small 0.30 \\  
         & & \small{$\textsc{BART}$} & \small{$\textsc{ExtAbs}(\textsc{BART})$} & \small ROUGE-L F1 & \small 0.53 & \small 0.30 \\ 
         & & \small{$\textsc{ExtAbs}(\textsc{PEGASUS})$} & \small{$\textsc{PEGASUS}$} & \small ROUGE-1 F1 & \small 0.68 & \small 0.25 \\ 
         & & \small{$\textsc{PEGASUS}$} & \small{$\textsc{ExtAbs}(\textsc{PEGASUS})$} & \small ROUGE-2 F1 & \small 2.36 & \small \textbf{0.01} \\  
         & & \small{$\textsc{PEGASUS}$} & \small{$\textsc{ExtAbs}(\textsc{PEGASUS})$} & \small ROUGE-L F1 & \small 0.95 & \small 0.17 \\ 
        \hline
        \multirow{12}{*}{\small \textbf{PubMed}} & \multirow{6}{*}{\small Extractive} & \small{$\textsc{ExtAbs}(\textsc{BART})$} & \small{$\textsc{MatchSum}$} & \small ROUGE-1 F1 & \small 29.43 & \small \textbf{5.88e-176} \\
         & & \small{$\textsc{ExtAbs}(\textsc{BART})$} & \small{$\textsc{MatchSum}$} & \small ROUGE-2 F1 & \small 28.15 & \small \textbf{3.36e-162} \\  
         & & \small{$\textsc{ExtAbs}(\textsc{BART})$} & \small{$\textsc{MatchSum}$} & \small ROUGE-L F1 & \small 45.60 & \small \textbf{0.00} \\ 
         & & \small{$\textsc{ExtAbs}(\textsc{PEGASUS})$} & \small{$\textsc{MatchSum}$} & \small ROUGE-1 F1 & \small 31.84 & \small \textbf{6.47e-203} \\ 
         & & \small{$\textsc{ExtAbs}(\textsc{PEGASUS})$} & \small{$\textsc{MatchSum}$} & \small ROUGE-2 F1 & \small 29.51 & \small \textbf{8.07e-177} \\
         & & \small{$\textsc{ExtAbs}(\textsc{PEGASUS})$} & \small{$\textsc{MatchSum}$} & \small ROUGE-L F1 & \small 47.87 & \small \textbf{0.00} \\ 
        \cdashline{2-7}
         & \multirow{6}{*}{\small Abstractive} & \small{$\textsc{ExtAbs}(\textsc{BART})$} & \small{$\textsc{BART}$} & \small ROUGE-1 F1 & \small 3.79 & \small \textbf{7.56e-05} \\ 
         & & \small{$\textsc{BART}$} & \small{$\textsc{ExtAbs}(\textsc{BART})$} & \small ROUGE-2 F1 & \small 4.93 & \small \textbf{4.27e-07} \\  
         & & \small{$\textsc{ExtAbs}(\textsc{BART})$} & \small{$\textsc{BART}$} & \small ROUGE-L F1 & \small 3.86 & \small \textbf{5.75e-05} \\ 
         & & \small{$\textsc{ExtAbs}(\textsc{PEGASUS})$} & \small{$\textsc{PEGASUS}$} & \small ROUGE-1 F1 & \small 1.97 & \small \textbf{0.02} \\ 
         & & \small{$\textsc{ExtAbs}(\textsc{PEGASUS})$} & \small{$\textsc{PEGASUS}$} & \small ROUGE-2 F1 & \small 1.78 & \small \textbf{0.04} \\  
         & & \small{$\textsc{ExtAbs}(\textsc{PEGASUS})$} & \small{$\textsc{PEGASUS}$} & \small ROUGE-L F1 & \small 2.45 & \small \textbf{0.01} \\ 
        \hline
    \end{tabular}
    \caption{Results of z-test. p-value is bold if it is less than 0.05, indicating the \textbf{statistical significance}.}
    \label{Tab:appendix_significance}
\end{table*}

\section{Discussion}
\label{appendix:discussion}
\begin{figure*}
      \centering
      \includegraphics[scale=0.5]{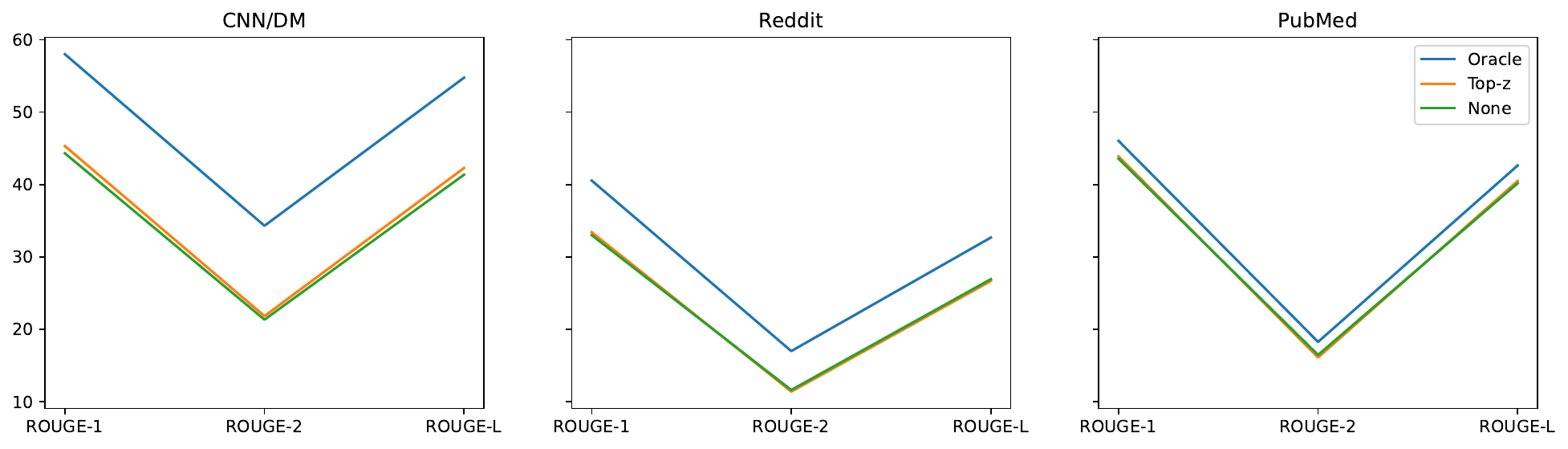}
      \caption{Visualization of ROUGE scores on $\textsc{ExtAbs}(\textsc{BART})$ with and without saliency mask across three datasets. \textcolor{c3}{Oracle} and \textcolor{c1}{Top-$z$} refer to saliency masks derived from the reference summary and top-$z$ textual segments predicted by the extractor, respectively. \textcolor{c2}{None} indicates the vanilla model $\textsc{BART}$.}
      \label{Fig:error_analysis}
\end{figure*}

We further investigate the inconsistent model improvements across the three datasets in abstractive summarization.
To demonstrate that a more accurate saliency mask contributes to a better abstractive summary and to quantify the potential boost from the saliency mask, we compare the ROUGE scores of abstractive summaries generated by $\textsc{ExtAbs}(\textsc{BART})$ using the oracle saliency mask (derived from the reference summary) versus the top-$z$ saliency mask (derived from the scores predicted by the extractor). Figure \ref{Fig:error_analysis} presents the results. The score difference between the oracle saliency mask and no saliency mask serves as the upper bound of potential boost from the saliency mask. 

Compared to CNN/DM, which has relatively high extractive scores and a higher potential boost for abstractive summaries, the model performance on Reddit is constrained by lower extractive scores (relatively less accurate saliency mask despite being SOTA). Examples in Table \ref{Tab:appendix_cnndm_example} and Table \ref{Tab:appendix_reddit_example} highlight the high extractive summary quality for CNN/DM and the lower quality for Reddit. PubMed, on the other hand, is limited by a lower upper bound, meaning the saliency mask provides less improvement compared to the other two datasets. A potential reason is the large number of extractive segments forming the saliency mask, reducing the difference between the saliency mask and the original non-padded mask (where all tokens are treated as salient). Specifically, the average number of EDUs in the input documents for the training sets of CNN/DM, Reddit, and PubMed is 94, 65 and 50, respectively, while the average number of EDUs used for saliency masks is 7, 5 and 22, respectively. 

\begin{table*}[ht!]
    \centering
    \begin{tabular}{p{0.11\textwidth}p{0.83\textwidth}}
        \hline
         {\small \textit{Document}} & {\small Have you ever panicked about whether you locked the front door, been woken by a Facebook notification or desperately needed a phone charger? To deal with these sorts of everyday emergencies, one company has designed a range of prototype products. \textit{\textcolor{c4}{You can vote on which one is your favourite - and the winner will be} made into an actual product, \textcolor{c4}{funded through a Kickstarter campaign.}} \textit{\textcolor{c1}{Leeds-based Direct Line is running a competition called \#EverydayFix.}} \textit{\textcolor{c2}{They asked groups to design products to deal with common problems.}} \textit{\textcolor{c3}{These include forgetting to lock the door,}} which one company hopes to solve with their Forget Me Lock prototype, shown. (...) 1. Not being able to sleep (46 per cent*) 2. Losing your keys (37 per cent) 3. Being stuck in traffic when already late (35 per cent) 4. Losing an important paper or document (33 per cent) 5. Nowhere to park (32 per cent) 6. Printer not working when you need to print something (31 per cent) 7. \textcolor{c3}{Running out of battery} on your phone whilst out (31 per cent) 8. Discovering you are out of toilet roll whilst on the loo (30 per cent) 9. Dealing with machine operated customer service (26 per cent) 10. Forgetting your bank card when paying for an item (25 per cent) (...) Another group designed Nipper - the world’s smallest mobile phone charger according to its creators, Impulse. (...) \textit{Direct Line is running a competition on Twitter} where you can vote for your favourite of these three products by retweeting tweets about Snooze, Forget Me Lock or Nipper. (...)} \\
         \hline
         {\small \textit{Reference Summary}} & {\small \textcolor{c1}{Leeds-based direct line is running a competition called \#everydayfix.} \textcolor{c2}{They asked groups to design products to deal with common problems.} \textcolor{c3}{These include forgetting to lock the door and running out of battery.} \textcolor{c4}{You can vote for your favourite design and the winner will be crowdfunded.}} \\
         \hline
         {\small $\textsc{BART}$} & {\small \textcolor{c1}{Leeds-based direct line is running a competition called \#everydayfix.} \textcolor{c2}{They asked groups to design products to deal with common problems.} \textcolor{c3}{These include forgetting to lock the door,} \textcolor{gray}{which one company hopes to solve with their forget me lock prototype. Another group designed Nipper - the world's smallest mobile phone charger.} \textcolor{c4}{You can vote on which one is your favourite - and the winner will be} \textcolor{gray}{made into an actual product,} \textcolor{c4}{funded through a Kickstarter campaign.}} \\
         \hline
         {\small $\textsc{ExtAbs}$ $(\textsc{BART})$-ext} & {\small \textcolor{c4}{You can vote on which one is your favourite - and the winner will be} \textcolor{gray}{made into an actual product,} \textcolor{c4}{funded through a Kickstarter campaign.} \textcolor{c1}{Leeds-based direct line is running a competition called \#everydayfix.} \textcolor{c2}{They asked groups to design products to deal with common problems.} \textcolor{gray}{Direct Line is running a competition on Twitter.}} \\
         \hline
         {\small $\textsc{ExtAbs}$ $(\textsc{BART})$-abs} & {\small \textcolor{c1}{Leeds-based direct line is running a competition called \#everydayfix.} \textcolor{c2}{They asked groups to design products to deal with common problems.} \textcolor{c3}{These include forgetting to lock the door,} \textcolor{gray}{forgetting to turn off the lights and forgetting to change the locks.} \textcolor{c4}{You can vote on which one is your favourite - and the winner will be funded through a Kickstarter campaign.}} \\
         \hline
    \end{tabular}
    \caption{CNN/DM summary example from our proposed model. Each colour represents a piece of salient information. \textit{The italic words} in the document are the salient tokens for the saliency mask.}
    \label{Tab:appendix_cnndm_example}
\end{table*}

\begin{table*}[ht!]
    \centering
    \begin{tabular}{p{0.11\textwidth}p{0.83\textwidth}}
        \hline
         {\small \textit{Document}} & {\small So this fuck up was yesterday and I am still starstruck from the incident. The day was finally over and I was so excited to leave and go home to play. \textit{I don't usually walk home from school} because it takes almost an hour to walk from my school to my house but my mom called saying she was gonna get off of work late \textcolor{c1}{\textit{and I was better off just walking home.}} All was going well until I decided to take out my phone and see what my friends were chatting about. \textcolor{c1}{\textit{While looking at my phone I didn't notice a tree root growing in the middle of the side walk and tripped over it.}} \textcolor{c2}{I hit the ground pretty hard and was starting to bleed from my arm, the worst part was that my bag ripped} and all my books and papers were all over the floor. I wasn't even half way to my house \textit{and \textcolor{c3}{I had to carry had to carry my books} in my hands along with my ripped up backpack with a bleeding arm.} granted, these books are almost 10 pounds and I had to carry 5 of them. \textit{I finally got home} and I realized \textcolor{c4}{I didn't even bring my keys to school that day.} \textit{I just left all my school supplies in front of my house,} went to my nearest Walmart and bought some bandages and disinfectant spray for my wound. I left Walmart went to go get something to eat and \textcolor{c5}{I still had another 4 hours until my mom came home.}} \\
         \hline
         {\small \textit{Reference Summary}} & {\small \textcolor{c1}{While walking home I decided to check my phone and tripped,} \textcolor{c2}{ripped my bag and scraped my arm.} \textcolor{c3}{Had to carry all my books,} \textcolor{c4}{forgot my keys} \textcolor{c5}{and had to wait in front of my house for 4 hours.}} \\
         \hline
         {\small $\textsc{BART}$} & {\small \textcolor{c1}{I tripped over a tree root on my way home from school,} \textcolor{c2}{ripped my backpack and ripped my arm,} \textcolor{c4}{didn't bring my keys}} \\
         \hline
         {\small $\textsc{ExtAbs}$ $(\textsc{BART})$-ext} & {\small \textcolor{gray}{I don't usually walk home from school} \textcolor{c1}{and I was better off just walking home.} \textcolor{c1}{While looking at my phone I didn't notice a tree root growing in the middle of the side walk and tripped over it.} \textcolor{c3}{and I had to carry had to carry my books} \textcolor{gray}{in my hands along with my ripped up backpack with a bleeding arm.} \textcolor{gray}{I finally got home I just left all my school supplies in front of my house,}} \\
         \hline
         {\small $\textsc{ExtAbs}$ $(\textsc{BART})$-abs} & {\small \textcolor{c1}{Tried to walk home from school, tripped over a tree root,} \textcolor{c2}{broke my arm,} \textcolor{c3}{and had to carry my school supplies to my house}} \\
         \hline
    \end{tabular}
    \caption{Reddit summary example from our proposed model. Each colour represents a piece of salient information. \textit{The italic words} in the document are the salient tokens for the saliency mask.}
    \label{Tab:appendix_reddit_example}
\end{table*}

\section{GPT-4 Evaluation}
\label{Appendix:gpt_evaluation}
We also conduct evaluation via $\textsc{GPT-4}$ to cover more evaluation metrics, including the coherence, fluency, consistency and relevance of the generated summary. Following the reason-then-score evaluation prompts designed by \citet{shen-etal-2023-large}, the "gpt-4-turbo" model is asked to rate the summary's coherence, fluency, consistency and relevance on a 5-point Likert scale where 5 means the best. Table \ref{Tab:gpt4_eval_prompts} lists the prompts and Table \ref{Tab:gpt4_eval} reports the averaged scores for each dimension. The generated abstractive summaries are scored only slightly lower on all dimensions than the reference summaries. In contrast, the extractive summaries gain much lower scores on dimensions of coherence and fluency but the highest score on consistency. The lower coherence and the higher consistency can both be explained by the nature of extractive summarization, which ignores the connection between extracted segments but is faithful to the original document. The relatively lower fluency score is due to the granularity of extracted textual segments. 

\begin{table*}[ht!]
    \centering
    \begin{tabular}{p{0.18\textwidth}p{0.78\textwidth}}
        \hline
        \textbf{\small Aspect} & \textbf{\small Prompt} \\
        \hline
         {\small Relevance} & {\small Score the following Summary given the corresponding Article with respect to relevance from one to five, where one indicates "irrelevance", and five indicates "perfect relevance". Note that relevance measures the Summary's selection of important content from the Article, whether the Summary grasps the main message of the Article without being overwhelmed by unnecessary or less significant details. Article: \textit{<article>}. Summary: \textit{<summary>}. Provide your reason in one sentence, then give a final score: } \\
         \hline
         {\small Consistency} & {\small Score the following Summary given the corresponding Article with respect to consistency from one to five, where one indicates "inconsistency" and five indicates "perfect consistency". Note that consistency measures the factual alignment between the Summary and the Article, whether the Summary is faithful to the Article without introducing contradictions or misleading representations. Article: \textit{<article>}. Summary: \textit{<summary>}. Provide your reason in one sentence, then give a final score: } \\
         \hline
         {\small Fluency} & {\small Score the following Summary given the corresponding Article with respect to fluency from one to five, where one indicates "disfluency" and five indicates "perfect fluency". Note that fluency measures the quality of individual sentences in the Summary, whether the Summary is well-written, grammatically correct, and readable on the sentence level. Article: \textit{<article>}. Summary: \textit{<summary>}. Provide your reason in one sentence, then give a final score: } \\
         \hline
         {\small Coherence} & {\small Score the following Summary given the corresponding Article with respect to coherence from one to five, where one indicates "incoherence" and five indicates "perfect coherence". Note that coherence measures the collective quality of the Summary, whether the Summary presents information that flows smoothly and avoids abrupt transitions or disjoint statements. Article: \textit{<article>}. Summary: \textit{<summary>}. Provide your reason in one sentence, then give a final score: } \\
         \hline
    \end{tabular}
    \caption{Prompts for $\textsc{GPT-4}$ evaluator.}
    \label{Tab:gpt4_eval_prompts}
\end{table*}

\begin{table*}[ht!]
    \centering
    \begin{tabular}{lcccc}
        \hline
         \textbf{Model} & \textbf{Coherence} & \textbf{Fluency} & \textbf{Consistency} & \textbf{Relevance} \\
         \hline
         Reference summary & 4.63 & 4.77 & 4.73 & 4.73 \\
         $\textsc{ExtAbs}(\textsc{BART})$-ext & 3.63 & 4.20 & 4.90 & 4.60 \\
         $\textsc{ExtAbs}(\textsc{BART})$-abs & 4.53 & 4.70 & 4.50 & 4.70 \\
         \hline
    \end{tabular}
    \caption{Evaluation results by the $\textsc{GPT-4}$ evaluator.}
    \label{Tab:gpt4_eval}
\end{table*}

\end{document}